\RequirePackage[l2tabu, orthodox]{nag}
\documentclass[11pt]{article}

\usepackage[T1]{fontenc}
\usepackage{tcolorbox}

\usepackage{soul} 

\usepackage{garamond}  
\renewcommand{\rmdefault}{ugm}

\usepackage[bitstream-charter]{mathdesign}
\usepackage{amsmath}
\usepackage[scaled=0.92]{PTSans}

\usepackage[
  paper  = letterpaper,
  left   = 1.65in,
  right  = 1.65in,
  top    = 1.0in,
  bottom = 1.0in,
  ]{geometry}

\definecolor{shadecolor}{gray}{0.9}

\usepackage[final,expansion=alltext]{microtype}
\usepackage[english]{babel}
\usepackage[parfill]{parskip}
\usepackage{afterpage}
\usepackage{framed}

{\endMakeFramed}

\DeclareRobustCommand{\parhead}[1]{\textbf{#1}~}

\usepackage{lineno}

\usepackage{ragged2e}

\newcounter{parcount}

\usepackage{graphicx}
\usepackage[labelfont=bf]{caption}

\usepackage{booktabs}
\usepackage{multirow}

\usepackage[algoruled]{algorithm2e}
\usepackage{listings}
\usepackage{fancyvrb}
\fvset{fontsize=\normalsize}

\usepackage{natbib}

\usepackage[acronym,nowarn]{glossaries}
\makeglossaries

\definecolor{tangerine}{rgb}{0.95, 0.52, 0.0}
\definecolor{palebrown}{rgb}{0.6, 0.46, 0.33}
\definecolor{peru}{rgb}{0.8, 0.52, 0.25}
\usepackage[colorlinks,linktoc=all]{hyperref}
\usepackage[all]{hypcap}
\hypersetup{citecolor=peru}
\hypersetup{linkcolor=peru}
\hypersetup{urlcolor=peru}

\usepackage[nameinlink]{cleveref}
\creflabelformat{equation}{#1#2#3}
\crefname{equation}{eq.}{eqs.}  
\Crefname{equation}{Eq.}{Eqs.}

\lstdefinestyle{mystyle}{
    commentstyle=\color{OliveGreen},
    keywordstyle=\color{BurntOrange},
    numberstyle=\tiny\color{black!60},
    stringstyle=\color{MidnightBlue},
    basicstyle=\ttfamily,
    breakatwhitespace=false,
    breaklines=true,
    captionpos=b,
    keepspaces=true,
    numbers=left,
    numbersep=5pt,
    showspaces=false,
    showstringspaces=false,
    showtabs=false,
    tabsize=2
}
\usepackage[colorinlistoftodos,
           prependcaption,
           textsize=small,
           backgroundcolor=yellow,
           linecolor=lightgray,
           bordercolor=lightgray]{todonotes}

\DeclareRobustCommand{\parhead}[1]{\textbf{#1}~}

\usepackage{amsthm}  

\usepackage{bm}

\usepackage[colorinlistoftodos,
           prependcaption,
           textsize=small,
           backgroundcolor=yellow,
           linecolor=lightgray,
           bordercolor=lightgray]{todonotes}

\usepackage{graphicx}
\usepackage{caption}
\usepackage{subcaption}
\usepackage{wrapfig}

\usepackage{booktabs}
\usepackage{arydshln} 
\usepackage{multirow}
\usepackage{nicematrix}

\usepackage{listings}
\usepackage{fancyvrb}
\fvset{fontsize=\normalsize}

\usepackage[nameinlink]{cleveref}
\creflabelformat{equation}{#1#2#3}
\crefname{equation}{eq.}{eqs.}  
\Crefname{equation}{Eq.}{Eqs.}

\usepackage{listings}
\lstdefinestyle{alp_style}{
    commentstyle=\color{OliveGreen},
    numberstyle=\tiny\color{black!60},
    stringstyle=\color{BrickRed},
    basicstyle=\ttfamily\scriptsize,
    breakatwhitespace=false,
    breaklines=true,
    captionpos=b,
    keepspaces=true,
    numbers=none,
    numbersep=5pt,
    showspaces=false,
    showstringspaces=false,
    showtabs=false,
    tabsize=2
}

\usepackage{capt-of}

\usepackage{lipsum}

\theoremstyle{remark}
\newtheorem*{lemma*}{Lemma}

\newcommand{\bz}{\bm{z}}

\newcommand{\bx}{\bm{x}}

\newcommand{\bmu}{\bm{\mu}}

\usepackage{authblk}
\usepackage{enumitem} 
\usepackage{algorithmic}
\usepackage{tabularx}
\usepackage{booktabs}
\usepackage{pifont}
\usepackage[T1]{fontenc}
\usepackage{multirow}
\usepackage{threeparttable}

\usetikzlibrary{calc}
\usepackage{tikz}
\usepackage{amsmath}
\usetikzlibrary{arrows.meta, positioning, calc, decorations.pathreplacing}
\usepackage{adjustbox}

\title{\textbf{Alternators With Noise Models}}

\author[2]{Mohammad R. Rezaei}
\author[1, 2]{Adji Bousso Dieng}
\affil[1]{Department of Computer Science, Princeton University}
\affil[2]{\href{https://vertaix.princeton.edu/}{Vertaix}}

\makeglossary 

\begin{document}
\maketitle

\begin{abstract}
\noindent Alternators have recently been introduced as a framework for modeling time-dependent data. They often outperform other popular frameworks, such as state-space models and diffusion models, on challenging time-series tasks. This paper introduces a new Alternator model, called \textbf{Alternator++}, which enhances the flexibility of traditional Alternators by explicitly modeling the noise terms used to sample the latent and observed trajectories, drawing on the idea of noise models from the diffusion modeling literature. Alternator++ optimizes the sum of the Alternator loss and a noise-matching loss. The latter forces the noise trajectories generated by the two noise models to approximate the noise trajectories that produce the observed and latent trajectories. We demonstrate the effectiveness of Alternator++ in tasks such as density estimation, time series imputation, and forecasting, showing that it outperforms several strong baselines, including Mambas, ScoreGrad, and Dyffusion.\\

\noindent \textbf{Keywords:} Time-Series, Dynamics, Latent Variables, Diffusion, Dynamical Systems, Imputation, Forecasting, Alternators, Machine Learning 
\end{abstract}

\section{Introduction}
Modeling complex time-dependent data is a central challenge in science and engineering. Recent advancements in sequence modeling are based on two popular frameworks: structured state-space models (SSMs) such as Mamba~\citep{gu2023mamba} and diffusion models~\citep{ho2020denoising, rasul2021autoregressive}. These approaches have been successfully applied across various domains, including natural language processing~\citep{gu2023mamba}, computer vision~\citep{zhu2024vision, rombach2022high}, and computational biology~\citep{xu2024protein}. They provide powerful tools for sequence modeling by capturing complex dependencies and offering strong generative capabilities.

Despite these successes, SSMs and diffusion models face significant challenges. They employ hidden representations that have the same dimensionality as the data, which leads to large models with high computational training costs. Furthermore, Mamba struggles with capturing long-range dependencies in noisy signals due to its reliance on structured state transitions in its network architecture~\citep{wang2025mamba}. These state transitions can be affected by noise that propagates through time, which can be limiting when processing highly noisy time-series~\citep{wang2025mamba, rezaei2025alpha}. Diffusion models, on the other hand, are notably slow to generate new data from, with significant research dedicated to accelerating their sampling process~\citep{song2020denoising, vahdat2021score, salimans2022progressive, lu2022dpm, karras2022elucidating}.

Alternators have been recently introduced as an alternative framework for sequence modeling~\citep{rezaei2024alternators}. They offer a more efficient latent representation by maintaining a low-dimensional state space, reducing computational complexity while preserving expressivity. However, Alternators assume a fixed noise distribution when sampling observation and latent trajectories, which may be limiting. 

In this paper, we introduce Alternator++, a new member of the Alternator class of models that uses trainable noise models instead of fixed probability distributions to define the noise terms used to generate observation and latent trajectories. Noise models have proven to be very beneficial for diffusion models~\citep{dhariwal2021diffusion, ho2022imagen, nichol2021improved}; they improve the quality of the generated outputs~\citep{ho2020denoising, song2020denoising}, enable stable training~\citep{lin2024common, chen2023importance}, and enhance model robustness~\citep{lee2024ant}. Alternator++ inherits these advantages while efficiently generating observation and latent trajectories following the Alternator framework. More specifically, while the noise terms in the original Alternator had zero means, leveraging noise models lifts that restriction and allows us to learn the mean of the noise variables instead. These means are modeled using two neural networks, which are trained by adding a noise-matching objective in the Alternator loss. 

Through comprehensive experiments across multiple datasets and domains, we demonstrate that Alternator++ consistently outperforms Mamba, diffusion models, and the original Alternator on density estimation, time-series imputation, and forecasting. 

\section{Background}
\label{sec:background}

Here we provide background on the two foundations of Alternator++: Alternators and noise models. 

\subsection{Diffusion Models}
Diffusion models are a powerful approach to generative modeling. The framework consists of two processes: the forward (diffusion) process progressively adds noise to the observations, while the reverse (denoising) process removes the noise from the observations.

\parhead{Forward diffusion and reverse denoising processes.} Let $\bx_0 \in \mathbb{R}^{D_x}$ be a data point. The forward diffusion process of a diffusion model is a Markov chain which iteratively adds Gaussian noise to $\bx_0$ until, after $T$ iteration steps, the observation at that time step, denoted by $\bx_T$, is nearly a sample from a standard Gaussian. Concretely, for a fixed schedule $\{\beta_t \in (0,1)\}_{t=1}^T$, the transition from one step to the next is characterized by the conditional distribution
\begin{align}
    q(\bx_t \mid \bx_{t-1}) &= \mathcal{N}\bigl(\bx_t ; \sqrt{1-\beta_t}\,\bx_{t-1},\,\beta_t \mathbf{I}\bigr),
\end{align}
such that $\bx_t$ is a noised version of $\bx_{t-1}$. By defining $\alpha_t = \prod_{s=1}^t (1-\beta_s)$, one can directly relate $\bx_t$ to $\bx_0$ through the conditional distribution
\begin{align}
    q(\bx_t \mid \bx_0) &= \mathcal{N}\bigl(\bx_t ; \sqrt{\alpha_t}\,\bx_0,
(1-\alpha_t)\mathbf{I}\bigr).
\end{align}

The reverse denoising process is characterized by the conditional distribution 
\begin{align}
    p_\theta(\bx_{t-1}\mid \bx_t) = \mathcal{N}\left(\bx_{t-1}; \frac{1}{\sqrt{1-\beta_t}} \left(\bx_t - \frac{\beta_t}{\sqrt{1-\alpha_t}}\, \epsilon_\theta(\bx_t, t)\right), \beta_t \mathbf{I}\right)
\end{align}
where $\epsilon_\theta$ is a neural network, called a \emph{noise model}, that takes $\bx_t$ and $t$ as input.

\parhead{Learning.} The parameters $\theta$ described above are learned via denoising score matching. Specifically, one trains the neural network $\epsilon_\theta$ to predict the noise $\epsilon$ that was added at step $t$ by minimizing 
\begin{align}
    \mathcal{L}_{\text{diff}} &= \mathbb{E}_{t,\bx_0,\epsilon} \bigl[ \|\epsilon - \epsilon_\theta(\bx_t, t)\|_2^2 \bigr]
\end{align}
where $\epsilon \sim \mathcal{N}(\mathbf{0}, \mathbf{I})$ and $\bx_t = \sqrt{\alpha_t}\,\bx_0 + \sqrt{1-\alpha_t}\,\epsilon$. Minimizing this objective makes $\epsilon_\theta$ an effective denoiser. Equivalently, $\epsilon_\theta$ approximates the score function $\nabla_{\bx_t} \log p(\bx_t)$ (up to a scaling factor) at each time $t$.

\parhead{Sampling.} Once trained, the reverse denoising process can be approximated by a discretized Langevin dynamics update:
\begin{align}
\bx_{t-1} &= \frac{1}{\sqrt{1-\beta_t}} \Bigl(\bx_t - \frac{\beta_t}{\sqrt{1-\alpha_t}}\, \nabla_{\bx_t}\log p(\bx_t)\Bigr) + \sqrt{\beta_t}\,\boldsymbol{\epsilon},
\end{align}
where $\boldsymbol{\epsilon} \sim \mathcal{N}(\mathbf{0}, \mathbf{I})$ and $\nabla_{\bx_t}\log p(\bx_t)$ is replaced by the neural network's score estimate. Each step removes a small amount of noise and adds a controlled Gaussian perturbation, ultimately yielding a fully denoised generated sample $\bx_0$.

\subsection{Alternators}
Consider a sequence $\bx_{1:T}$. An Alternator models this sequence by pairing it with latent variables $\bz_{0:T}$ in a joint distribution \cite{rezaei2024alternators}:
\begin{align}\label{eq:joint}
    p_{\theta, \phi}(\bx_{1:T}, \bz_{0:T}) &= p(\bz_0)\, \prod_{t=1}^{T} p_{\theta}(\bx_t \mid \bz_{t-1})\, p_{\phi}(\bz_t \mid \bz_{t-1}, \bx_t).
\end{align}
Here $p(\bz_0) = \mathcal{N}(0, \mathbf{I})$ is a prior over the initial latent variable $\bz_0$ and $p_{\phi}(\bz_t \mid \bz_{t-1}, \bx_t)$ models how the other latent variables are generated over time. The observations that it conditions on are modeled through $p_{\theta}(\bx_t \mid \bz_{t-1})$. Both conditional distributions are Gaussians parameterized by neural networks with parameters $\phi$ and $\theta$,
\begin{align*}
    p_{\theta}(\bx_t | \bz_{t-1}) &= \mathcal{N}\left(\bmu_{x_t},  \sigma_x^2\mathbf{I}\right) \text{ where }
    \bmu_{x_t} = \sqrt{(1 - \sigma_x^2)}\cdot f_{\theta}(\bz_{t-1})\\
    p_{\phi}(\bz_t | \bz_{t-1}, \bx_t) &= \mathcal{N}\left(\bmu_{z_t}, \sigma_z^2\mathbf{I}\right), \text{ where }
    \bmu_{z_t} = \sqrt{\alpha_t} \cdot g_{\phi}(\bx_t) + \sqrt{(1 - \alpha_t - \sigma_z^2)}\cdot \bz_{t-1}
\end{align*}
The parameters $\theta$ and $\phi$ are learned by minimizing the Alternator loss 
\begin{align}\label{eq:loss-alt}
    \mathcal{L}_{\text{Alternator}}(\theta, \phi) &= \mathbb{E}_{p(\bx_{1:T}) p_{\theta, \phi}(\bz_{0:T})}\left[\sum_{t=1}^{T} \left\Vert \bz_t - \bmu_{z_t} \right\Vert^2_2 + \frac{D_z \sigma_z^2}{D_x \sigma_x^2} \left\Vert \bx_t - \bmu_{x_t} \right\Vert^2_2 \right],
\end{align}
where $p(\bx_{1:T})$ is the data distribution and $p_{\theta, \phi}(\bz_{0:T})$ is the marginal distribution of the latent variables induced by the joint distribution in Eq.~\ref{eq:joint}. Alternators model sequences over time by coupling observations with latent variables, whereas diffusion models rely on iterative denoising. The Alternator++ model, introduced in the next section, extends the Alternator framework by incorporating a diffusion-based refinement step within the latent evolution process.

\section{Alternator++}
\label{sec:method}
We now describe the generative process of Alternator++ and the objective function used to train its parameters. 

\subsection{Generative Process}
\label{sec:generative-process}
While standard Alternators model a time-indexed sequence $\bx_{1:T}$ paired with latent variables $\bz_{0:T}$ using fixed noise distributions as shown in equation \ref{eq:joint}, Alternator++ uses trainable noise prediction networks $\boldsymbol{\epsilon^t_{\psi}}$ and $\boldsymbol{\epsilon^t_{\nu}}$ that flexibly model stochasticity at each time step t.

The generative process begins by sampling an initial latent variable $\bz_0 \sim \mathcal{N}(0, I_{D_z})$ from a standard Gaussian. Then we generate a sequence by alternating between generating observation $\bx_t$ conditioned on the previous latent state $\bz_{t-1}$ and updating the latent representation $\bz_t$ using the previous latent state $\bz_{t-1}$ and the current observation $\bx_t$. The key innovation in Alternator++ lies in its explicit noise modeling through the specialized networks $\boldsymbol{\epsilon^t_{\psi}}$ and $\boldsymbol{\epsilon^t_{\nu}}$ that dynamically adjust stochasticity levels when sampling the observed and latent trajectories. Indeed, for any $t$, we sample $\bx_t$ and $\bz_t$ as
\begin{align}
\label{eq:sample-x}
\bx_t &= \sqrt{\beta_t} \cdot f_{\theta}(\bz_{t-1}) + \sqrt{1 - \beta_t - \sigma_x^2} \cdot \boldsymbol{\epsilon^t_{\psi}}(\bz_{t-1}) + \sigma_x \boldsymbol{\epsilon_{\mu_{x_t}}}\\
\label{eq:sample-z}
\bz_t &= \sqrt{\alpha_t} \cdot g_{\phi}(\bx_t) + \sqrt{1 - \alpha_t - \sigma_z^2} \cdot \boldsymbol{\epsilon^t_{\nu}}(\bz_{t-1}, \bx_t) + \sigma_z \boldsymbol{\epsilon_{\mu_{z_t}}}
\end{align}
Here, $\boldsymbol{\epsilon_{\mu_{x_t}}} \sim \mathcal{N}(0, I_{D_x})$ and $\boldsymbol{\epsilon_{\mu_{z_t}}} \sim \mathcal{N}(0, I_{D_z})$ are standard Gaussian noise variables. The functions $f_{\theta}$ and $g_{\phi}$ map latent variables and observations, respectively, as in the original Alternator framework. They are both neural networks with parameters $\theta$ and $\phi$, respectively. The noise models $\boldsymbol{\epsilon^t_{\psi}}$ and $\boldsymbol{\epsilon^t_{\nu}}$ are modulated by time-dependent noise schedules $\beta_{1:T}$ and $\alpha_{1:T}$, with base variance parameters $\sigma_x^2$ and $\sigma_z^2$, respectively. 

The noise prediction network $\boldsymbol{\epsilon^t_{\nu}}$ takes both $\bz_{t-1}$ and $\bx_t$ as inputs to drive the dynamics of $\bz_t$, whereas the original Alternator used a simple interpolation of $\bz_{t-1}$ to update the latent $\bz_t$. Taking $\bx_t$ as an additional input adds more expressivity and makes the latent variables more context-aware. Another departure from the original Alternator is the network $\boldsymbol{\epsilon^t_{\psi}}$, which enhances the model’s ability to capture complex and time-varying noise patterns in the observation space. 

The noise schedules $\beta_{1:T}$ and $\alpha_{1:T}$ modulate the influence of the learned noise models. When $\beta_t \rightarrow 1 - \sigma_x^2$ and $\alpha_t \rightarrow 1 - \sigma_z^2$, the contributions of the noise prediction networks $\boldsymbol{\epsilon^t_{\psi}}$ and $\boldsymbol{\epsilon^t_{\nu}}$ diminish, and the generative dynamics revert to those of the original Alternator model. In contrast, as $\beta_t \rightarrow 0$, the generation of $\bx_t$ becomes increasingly influenced by the learned noise model $\boldsymbol{\epsilon^t_{\psi}}$. This allows the model to capture complex and time-varying noise patterns that are dependent on the latent state $\bz_{t-1}$, thus enabling a richer and more flexible description of stochasticity in the observation domain. Similarly, as $\alpha_t \rightarrow 0$, the noise model $\boldsymbol{\epsilon^t_{\nu}}$ jointly driven by current observation $\bx_t$ and previous latent variable $\bz_{t-1}$ has a greater influence on the prediction of the latent variable $\bz_t$.

\subsection{Training Objective}
The Alternator++ training objective adds a noise-matching objective to the original Alternator loss, 
\begin{align}
\label{eq:final_diffusion_loss}
{\mathcal{L}}(\theta, \phi)
&=  \mathcal{L}_{\text{alternator}}(\theta, \phi, \psi, \nu) +  \lambda \cdot \mathcal{L}_{\epsilon}(\theta, \phi, \psi, \nu)\\
\mathcal{L}_{\text{alternator}}(\theta, \phi, \psi, \nu) \nonumber
&= \frac{1}{B}\sum_{b=1}^{B}\sum_{t=1}^{T}
        \left[\left\Vert \bz_t^{(b)} - \bmu_{z_t^{(b)}} \right\Vert^2_2 + \frac{D_z \sigma_z^2}{D_x \sigma_x^2}\cdot \left\Vert \bx_t^{(b)} - \bmu_{x_t^{(b)}} \right\Vert^2_2\right]\\
\mathcal{L}_{\epsilon}(\theta, \phi, \psi, \nu) \nonumber
&=\frac{1}{B}\sum_{b=1}^{B}\sum_{t=1}^{T} \bigl\| \boldsymbol{\epsilon}_z^{(b)} - \boldsymbol{\epsilon^t_{\nu}}(\bz_{t-1}^{(b)}, \bx_t^{(b)}) \bigr\|_{2}^{2} + \gamma_t \cdot\bigl\| \boldsymbol{\epsilon}_x^{(b)} - \boldsymbol{\epsilon^t_{\psi}}(\bz_{t-1}^{(b)}) \bigr\|_{2}^{2}  
\end{align}
where $\bx_t^{(b)}$ is the $t^{\text{th}}$ observation of the $b^{\text{th}}$ sequence in the batch, it is drawn from the training data. On the other hand $\bz_t^{(b)}$ is the latent variable at time $t$ for the $b^{\text{th}}$ sequence in the batch, it is sampled using the generative process (\ref{eq:sample-x}, \ref{eq:sample-z}) with $\bz_0^{(b)} \sim \mathcal{N}(0, I_{D_z})$. The means $\bmu_{x_t}^{(b)} $ and $\bmu_{z_t}^{(b)}$ are defined as 
\begin{align}
    \bmu_{x_t}^{(b)} &= \sqrt{\beta_t} f_{\theta}^z(\bz_{t-1}^{(b)}) + \sqrt{1-\beta_t-\sigma_x^2} \cdot \boldsymbol{\epsilon^t_{\psi}}(\bz_{t-1}^{(b)})\\   
    \bmu_{z_t}^{(b)} &= \sqrt{\alpha_t} \cdot g_{\phi}^x(\bx_t^{(b)}) + \sqrt{1-\alpha_t-\sigma_z^2} \cdot \boldsymbol{\epsilon^t_{\nu}}(\bz_{t-1}^{(b)}, \bx_t^{(b)}).
\end{align}
The terms $\boldsymbol{\epsilon}_{x}^{(b)} \sim \mathcal{N}(0, I{D_x})$ and $\boldsymbol{\epsilon}_{z}^{(b)} \sim \mathcal{N}(0, I{D_z})$ are standard Gaussian noise variables sampled for each time step and batch element. Here $\gamma_t = \frac{D_z \sigma_z^2 \alpha_t}{D_x \sigma_x^2 \beta_t}$ balances the two noise-matching loss terms. This balancing prevents the model from prioritizing one space over the other simply due to differences in dimensionality or noise magnitude, ensuring consistent learning across both the observation and latent space noise models. Finally, $\lambda$ is a hyperparameter controlling the relative importance of noise prediction. When $\lambda$ is small, the model behaves more like the original Alternator, focusing on reconstruction accuracy. As $\lambda$ increases, the model places greater emphasis on learning accurate noise distributions, which improves its ability to model complex stochastic patterns. Algorithm~\ref{alg:training++} summarizes the training procedure for Alternator++.

\begin{algorithm}[t]
\DontPrintSemicolon
\textbf{Inputs}: Data $\bx_{1:T}^{(1:n)}$, batch size $B$, variances $\sigma_x^2$, $\sigma_z^2$, noise schedules $\beta_{1:T}$, $\alpha_{1:T}$\;
Initialize model parameters $\theta$, $\phi$, $\psi$, $\nu$\;
\While{not converged}{
    Sample a batch of sequences $\{\bx_{1:T}^{(b)}\}_{b=1}^B$ from the dataset\;
    \For{$b = 1, \dots, B$}{
        Draw initial latent $\bz_0^{(b)} \sim \mathcal{N}(0, I_{D_z})$\;
        \For{$t = 1, \dots, T$}{
            Draw noise samples $ \boldsymbol{\epsilon_{\mu_{z_t}}} \sim \mathcal{N}(0, I_{D_z})$ and $\boldsymbol{\epsilon_{\mu_{x_t}}} \sim \mathcal{N}(0, I_{D_x})$\;
            Compute $\bmu_{x_t}^{(b)} = \sqrt{\beta_t} \cdot f_{\theta}(\bz_{t-1}^{(b)}) + \sqrt{1-\beta_t-\sigma_x^2} \cdot \boldsymbol{\epsilon^t_{\psi}}(\bz_{t-1}^{(b)}) $\;
            Sample observation $\bx_{t}^{(b)} = \bmu_{x_t}^{(b)}+\sigma_x \boldsymbol{\epsilon_{\mu_{x_t}}}$\;
            Compute $\bmu_{z_t}^{(b)} = \sqrt{\alpha_t} \cdot g_{\phi}(\bx_t^{(b)}) + \sqrt{1-\alpha_t-\sigma_z^2} \cdot \boldsymbol{\epsilon^t_{\nu}}(\bz_{t-1}^{(b)}, \bx_t^{(b)}) $\;
            Sample latent $\bz_{t}^{(b)} = \bmu_{z_t}^{(b)}+\sigma_z \boldsymbol{\epsilon_{\mu_{z_t}}}$\;
        }
    }
    Compute loss $\mathcal{L}(\theta, \phi, \psi, \nu)$ using $(\bx_{1:T}, \bz_{0:T},\bmu_{z_{0:T}}, \bmu_{x_{1:T}})$\;
    Backpropagate and update parameters $\theta$, $\phi, \psi, \nu$ using the Adam optimizer
}
\caption{Sequence modeling with Alternator++}
\label{alg:training++}
\end{algorithm}

\subsection{Sampling and Encoding New Sequences}
After training, one can sample from Alternator++ to generate new sequences by simply using the generative process described in Section \ref{sec:generative-process}. That same generative process also indicates how to encode, i.e., get the low-dimensional representation, of a new sequence $\bx_{1:T}^*$: simply replace the sampled $\bx_t$ with the given $\bx^*_t$ and return $\bmu_{z_t}$ for $t \in \left\{1, \dots, T\right\}$. 

\section{Experiments}
\label{sec:empirical}

In this section, we present a comprehensive evaluation of Alternator++ across multiple time-series datasets and tasks. Our experiments aim to answer the following  questions:
\begin{itemize}
    \item How well does Alternator++ capture complex temporal dependencies and multimodal densities in time-series data compared to existing dynamical generative models? 
    \item Can Alternator++ effectively handle missing values and outperform state-of-the-art methods in time-series imputation?
    \item Does Alternator++ demonstrate superior forecasting accuracy, particularly in challenging real-world applications such as sea surface temperature prediction? 
\end{itemize}
To systematically address these questions, we compare Alternator++ against widely recognized baselines, including VRNN, SRNN, NODE-RNN, ScoreGrad, Mamba, and Dyffusion. Our results demonstrate that Alternator++ tends to outperform these baselines across multiple datasets. Notably, it captures time-series distributions better as evidenced by its lower maximum mean discrepancy (MMD) scores. Furthermore, Alternator++ can perform well at imputation even when the missing rate is very high. Finally, it also performs well at forecasting, while being significantly more computationally efficient. The following sections provide a detailed breakdown of these findings. For comprehensive details regarding implementation specifics and hyperparameter configurations across each experiment, we refer the reader to the Appendix~\ref{app:imp}.

\subsection{Density estimation}

\begin{table*}[t]
\centering
\caption{Alternator++ tends to outperform several strong baselines, and by a wide margin. Here, performance is measured in terms of the MMD between the distribution learned by each model and the ground truth distribution, using generated samples from the models and the data.}
\begin{tabular}{lcccc} 
 \hline
 Method  & Solar &Covid& Fred & NN5 \\ 
 \hline
 Alternator++ & \textbf{0.051$\pm$0.004}& 0.043$\pm$0.031& \textbf{0.039$\pm$0.005}& \textbf{0.088$\pm$0.008}\\
 Alternator & 0.123$\pm$0.002 &0.592$\pm$0.063& 0.281$\pm$0.002&0.310$\pm$0.002\\
 Mamba & 0.131$\pm$0.001&\textbf{0.025$\pm$0.052} & 0.185$\pm$0.003 & 0.253$\pm$0.021  \\
 ScoreGrad & 0.115$\pm$0.003 & 0.573$\pm$0.012 & 0.142$\pm$0.020 & 0.155$\pm$0.009\\
 VRNN & 0.848$\pm$0.005 & 1.106$\pm$0.002 & 1.328$\pm$0.005 & 2.109$\pm$0.001\\
 SRNN & 1.013$\pm$0.030 & 1.240$\pm$0.001 & 1.367$\pm$0.003 & 2.480$\pm$0.002 \\
 NODE-RNN & 0.132$\pm$0.013& 0.621$\pm$0.081 &0.479$\pm$0.127 & 0.427$\pm$0.103 \\
 \hline
\end{tabular}
\label{tab:Distribution-metrics}
\end{table*}

We benchmark Alternator++ against Alternators~\citep{rezaei2024alternators}, ScoreGrad~\citep{yan2021scoregrad}, Mamba~\citep{gu2023mamba}, VAE-based models (VRNN~\cite{chung2015recurrent}, SRNN~\cite{fraccaro2016sequential}), and Neural ODE-based models (NODE-RNN~\cite{chen2019latent}) in modeling the underlying probability distribution of time-series datasets. We use MMD to measure the goodness-of-fit between the generated samples and the ground truth distribution. Table~\ref{tab:Distribution-metrics} summarizes the results of this experiment. 

Alternator++ achieves the lowest MMD scores on three of the four datasets, outperforming the previous best method, ScoreGrad, by 66\% on Solar, 72\% on Fred, and 47\% on NN5. On the Covid dataset, Mamba exceeds Alternator++ by 42\%, albeit with greater variability. Among the baselines, ScoreGrad consistently beats Mamba—particularly on NN5 and Fred, where it reduces MMD by 40\% and 24\%, respectively—demonstrating its superior generalization across diverse time-series distributions.

\begin{figure*}[t]
    \includegraphics[width=\linewidth]{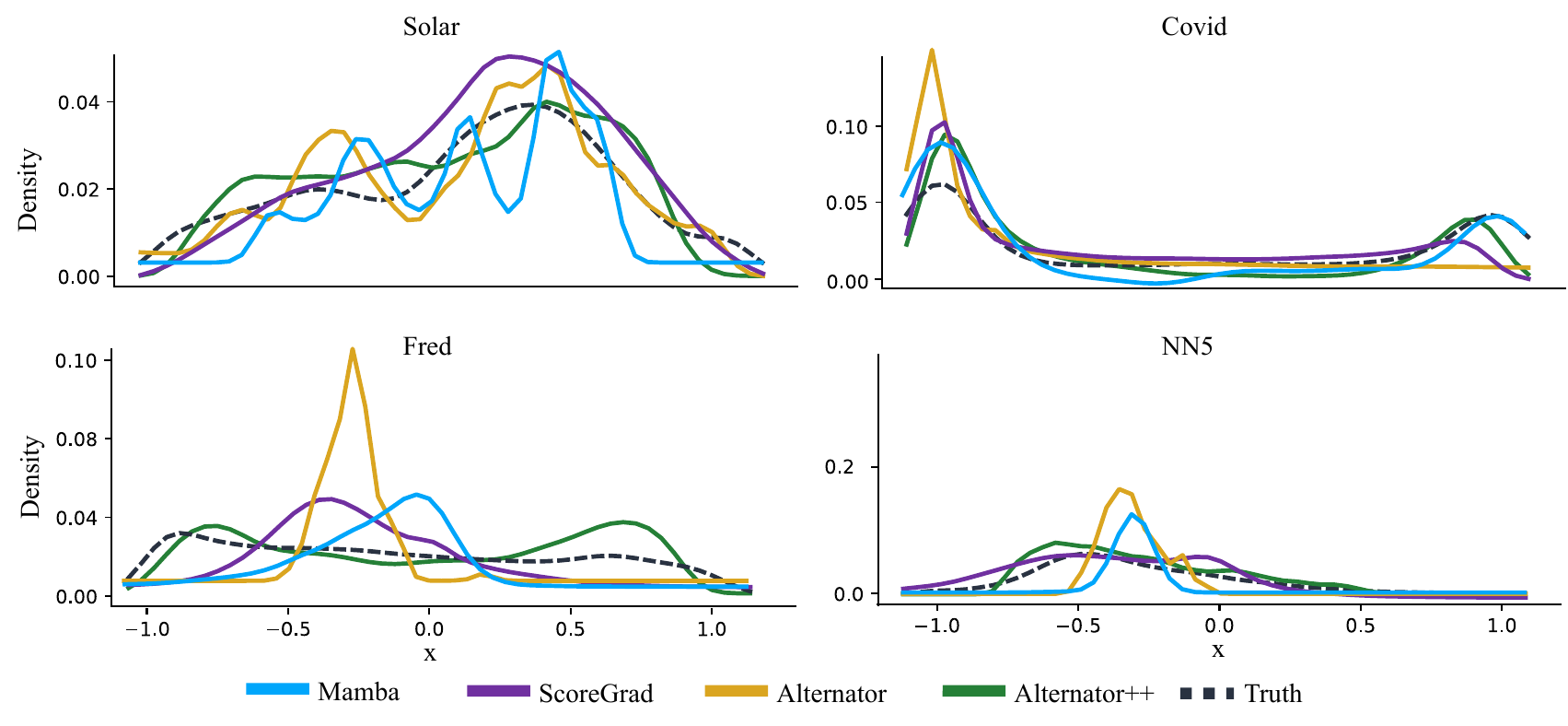}
    \caption{Comparing the distributions learned by various models against the ground truth distribution. Alternator++ captures multimodal distributions better than Alternator, Mamba, and ScoreGrad. 
    }
    \label{fig:Distributions}
\end{figure*}

These quantitative findings are corroborated in Figure \ref{fig:Distributions}. Alternator++’s samples align more closely with the target distribution on three out of four datasets. In highly skewed cases like Solar and NN5, it captures the distribution mode more accurately than any competitor. On the Covid and Fred datasets, Alternator++ correctly identifies both modes and assigns probability mass appropriately. The sole exception is the Covid dataset, where Mamba achieves better alignment.

\subsection{Time Series Imputation}
\begin{figure*}[t]
    \includegraphics[width=0.95\linewidth]{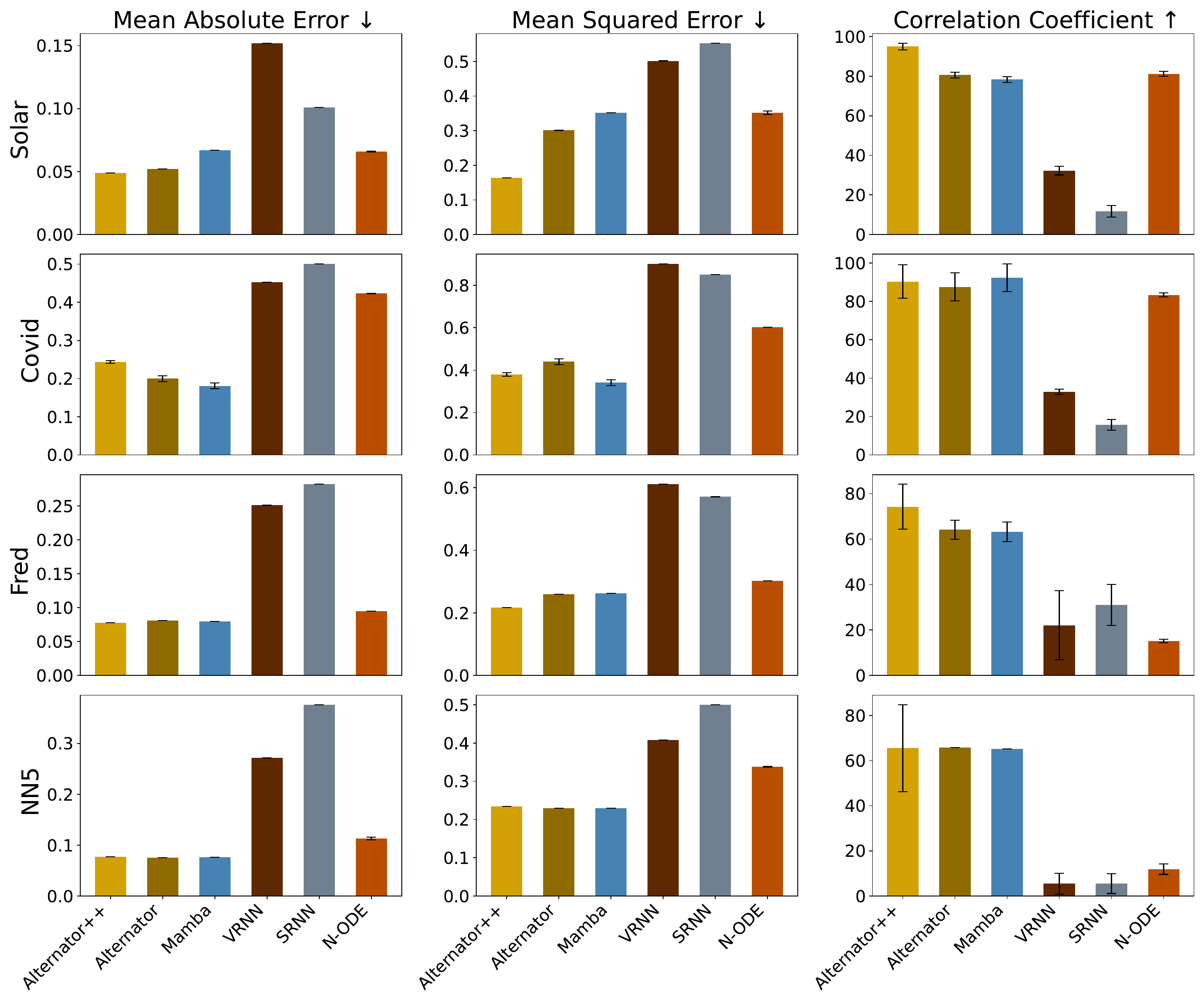}
    \caption{Performance on missing data imputation across several datasets, evaluated in terms of MAE, MSE, and CC. Results are averaged over missing rates ranging from 10\% to 90\%. Alternator++ generally outperforms the baselines in terms of MSE and CC. However, for MAE, it faces challenges on the Covid dataset, where Alternator and Mamba perform better.
    }
    \label{fig:Monash_imputation}
\end{figure*}

Time series imputation addresses scenarios where temporal observations contain missing values due to sensor malfunctions, transmission failures, or non-uniform sampling. We evaluate model robustness by varying the Missing At Random (MAR) rates from 10\% to 90\%. The results are summarized in Figure~\ref{fig:Monash_imputation}. Note, we did not use ScoreGrad for imputation because it uses diffusion-like processes optimized for unconditionally generating new samples from learned distributions. However, adapting this framework for imputation would require significant architectural modifications to the model investigated in the previous section to handle conditioning on partial observations. Additionally, imputation with ScoreGrad requires computationally expensive iterative sampling procedures per time step, which would be prohibitive for systematic evaluation across multiple missing rates from 10\% to 90\%. For these reasons, we excluded ScoreGrad from the imputation experiment.

On the Solar dataset, Alternator++ outperforms both Mamba and the original Alternator in mean absolute error (MAE), improving by over 20\% and 10\%, respectively. In mean squared error (MSE), Alternator++ reduces error by roughly 50\%, and its correlation coefficient is about 10\% higher. On the FRED dataset, Alternator++ again outperforms the baselines, achieving the lowest MAE, a substantially reduced MSE, and a correlation coefficient that exceeds competing methods by approximately 10\%. For NN5, Alternator++ maintains the best MAE, albeit with smaller margins, and consistently superior MSE and correlation. Finally, on the Covid dataset, Alternator++ outperforms the original Alternator in both MSE and correlation, though Mamba performs better on this dataset.

In summary, across Solar, FRED, and NN5, Alternator++ consistently achieves the lowest errors and highest correlations, demonstrating robust performance under varying patterns of missing data. Compared to the original Alternator, these results reflect clear gains in both accuracy and alignment with the true time series.

\subsection{Sea surface temperature forecasting}

\begin{table}[t]
  \caption{Performance on sea surface temperature forecasting with forecasting horizons ranging from 1 to 7 days ahead. Metrics are averaged over the entire evaluation horizon, with standard errors reported. For both CRPS and MSE, lower values indicate better performance. The time column indicates the total duration required to forecast all 7 future timesteps for a single batch. Alternator++ outperforms the baselines in terms of MSE and is relatively fast compared to MSDV and Dyffusion. However, it exhibits high standard errors and may underperform Mamba and Dyffusion in terms of CRPS.}
    \label{tab:sst-results}
    \centering
    \begin{tabular}{lccc}
    \toprule
   Method & CRPS & MSE  & Time [s] \\
    \midrule
    Perturbation & 
            0.281 $\pm$ 0.004 & 0.180 $\pm$ 0.011 & 0.4241 \\
    Dropout & 
            0.267 $\pm$ 0.003 & 0.164 $\pm$ 0.004 & 0.4241 \\
    DDPM & 
            0.246 $\pm$ 0.005 & 0.177 $\pm$ 0.005 & 0.3054 \\
    MCVD & 
            0.216 & 0.161 & 79.167 \\
    Dyffusion & 
            0.224 $\pm$ 0.001 & 0.173 $\pm$ 0.001 & 4.6722 \\
    Mamba & 
            0.219 $\pm$ 0.002 & 0.134 $\pm$ 0.003 & 0.6452 \\
    Alternator & 
            0.221 $\pm$ 0.031 & 0.144 $\pm$ 0.045 & 0.7524 \\
    Alternator++ & 
            \textbf{0.212 $\pm$ 0.040} & \textbf{0.116 $\pm$ 0.035} & 1.4277 \\
    \bottomrule
    \end{tabular}
\end{table}

In climate science, sea surface temperature (SST) prediction is crucial for weather forecasting and climate modeling ~\citep{haghbin2021applications}. We apply Alternator++ to forecast SST using a daily dataset from 1982-2021, with data split into training (1982-2019, 15,048 samples), validation (2020, 396 samples), and testing (2021, 396 samples). Following~\citep{cachay2023dyffusion}, we transform the global data into $60\times60$ (latitude$\times$longitude) tiles, selecting 11 patches in the eastern tropical Pacific Ocean for forecasting horizons of 1-7 days. 

We compare against Alternators~\citep{rezaei2024alternators}, DDPM~\citep{ho2020denoising}, MCVD~\citep{voleti2022mcvd}, DDPM variants (DDPM-D~\citep{gal2016dropout} and DDPM-P~\citep{pathak2022fourcastnet}), and Dyffusion~\citep{cachay2023dyffusion}, evaluating with CRPS~\citep{matheson1976scoring} and MSE, where CRPS is a proper scoring rule for probabilistic forecasting~\citep{gneiting2014probabilistic,de2020normalizing}, and MSE is measured on the mean prediction from a 50-member ensemble. 

Table~\ref{tab:sst-results} shows that Alternator++ achieves good performance on both metrics, improving CRPS by approximately $20\%$ over MCVD and reducing MSE by a similar margin compared to Dyffusion. While Alternator++ has slightly longer inference time than some baselines, it remains more than $50\times$ faster than MCVD and more than $3\times$ faster than Dyffusion for multi-lead time forecasts. 

\section{Related Work}
\label{sec:related-works}
The landscape of generative time-series modeling is rich with sophisticated approaches addressing the complex challenges posed by time-dependent data. Our work, Alternator++, builds upon and meaningfully distinguishes itself from several key paradigms that we discuss next.

\parhead{Neural ordinary differential equations} (N-ODEs), as explored by \citep{chen2018neural,rubanova2019latent}, provide differential equation solvers based on neural networks. However, their fundamentally deterministic nature is at odds with the stochasticity characterizing real-world time series data. Neural stochastic differential equations (Neural SDEs), introduced by \citet{liu2019neuralsdestabilizingneural}, address this by incorporating stochastic terms to model randomness. However, Neural SDEs typically rely on computationally expensive numerical solvers and maintain high-dimensional state representations. In contrast, Alternator++ maintains stochasticity through noise models for both latent and observation spaces that lean on more computationally efficient methods (score matching) compared to ODE/SDE solvers, while preserving expressiveness for complex temporal patterns.

\parhead{Variational recurrent neural networks} (VRNNs) marry RNNs with latent variables for sequential data modeling~\citep{fabius2014variational, fortunato2017bayesian,krishnan2015deep}. Fitting these models is often done using variational inference (VI). Different works have explored different ways of representing the variational distribution of the latent variables, with the richer variational distributions leveraging both past and future sequence elements for a given time step using bidirectional RNNs~\citep{bayer2014learning, fraccaro2016sequential,martinez2017human,doerr2018probabilistic,karl2016deep,castrejon2019improved}. However, they all face the challenge that at test time, the future isn't available, and sampling highly plausible sequences becomes difficult because of this. Alternator++ also relies on low-dimensional latent variables. However, instead of using VI for training, it uses the Alternator loss, which is a cross-entropy objective function on the observed and latent trajectories~\citep{rezaei2024alternators}. 

\parhead{State-space models} have proven effective across domains~\citep{gu2023mamba, rezaei2022direct, rezaei2021real, rangapuram2018deep}. Mamba~\citep{gu2023mamba} introduced selective SSMs, with subsequent domain-specific adaptations including Vision Mamba~\citep{zhu2024vision}, MambaStock~\citep{shi2024mambastock}, and protein models~\citep{xu2024protein}. Despite its versatility, Mamba uses a high-dimensional hidden state space ($\mathbf{h}_t \in \mathbb{R}^{d}$), making it computationally expensive, especially for long-horizon modeling. In addition to Mamba, other recent advances have significantly pushed the boundaries of state-space modeling by introducing long convolutions as an alternative to recurrence ~\citep{smith2023simplifiedstatespacelayers}, enabling subquadratic context length processing while maintaining competitive performance with transformers~\citep{gu2022efficientlymodelinglongsequences}. Alternator++ differs from these approaches by maintaining a low-dimensional latent state $\mathbf{z}_t \in \mathbb{R}^{d_z}$ where $d_z \ll d$ and employing trainable noise models, thus reducing complexity and improving generalization. 

\parhead{Diffusion models.} Alternator++ shares conceptual similarities with diffusion-based models like TimeGrad~\citep{rasul2021autoregressive}, Dyffusion~\citep{cachay2023dyffusion}, and others~\citep{karras2022elucidating, dhariwal2021diffusion,voleti2022mcvd, pathak2022fourcastnet,li2024constraint}. TimeGrad introduced diffusion for probabilistic forecasting, requiring hundreds of iterations to reconstruct signals. ScoreGrad~\citep{yan2021scoregrad} advanced this with continuous-time score-based frameworks, while Dyffusion~\citep{cachay2023dyffusion} incorporated physics-informed priors. Recent advances include CSDI~\citep{tashiro2021csdi}, DiffWave~\citep{kong2020diffwave}, TimeDiff~\citep{shen2023non}, DiffuSeq~\citep{gong2022diffuseq}, TDPM~\citep{ye2024schedule}, and ANT~\citep{lee2024ant}. Alternator++ differs in two main ways: (1) it uses noise-conditioned transitions, enabling direct next-step estimation without iterative perturbations, and (2) it models state transitions in a non-Markovian way, incorporating richer temporal relationships via learned noise components. 

\parhead{Alternators.} The original Alternator~\citep{rezaei2024alternators} employed a two-network architecture alternating between observation processing and latent state evolution. The $\alpha$-Alternator~\citep{rezaei2025alpha} dynamically adjusts the dependence on observations and latents when predicting an element of the sequence by using the Vendi Score~\citep{friedman2023vendi}, which makes it robust to varying noise levels in sequence data. Alternator++ is yet another extension of standard Alternators. It shifts from implicit to explicit noise modeling, using neural networks to model the means of the noise terms for both the observation and latent trajectories. 

\section{Conclusion}
\label{sec:conclusion}
We developed \textit{Alternator++}, a new Alternator model that leverages noise models from the diffusion modeling literature for improved performance. The noise models are neural networks whose parameters are learned by adding a noise-matching objective to the Alternator loss. By modeling the noise terms in both the latent and observed trajectories, Alternator++ captures complex temporal dynamics more accurately. We demonstrate this in experiments on density estimation, imputation, and forecasting tasks, where we found Alternator++ outperforms strong baselines such as Mamba, ScoreGrad, and Dyffusion. In addition to its generalization capabilities, Alternator++ offers fast sampling and low-dimensional latent variables, two features that diffusion models and state-space models lack. In combining low-dimensional latent representations with trainable noise models, Alternator++ enables both accurate modeling and computational efficiency. 

\parhead{Limitations}
Despite the promising results shown in this paper, Alternator++ can be extended to enhance performance even further. Indeed, the schedule parameters $\beta_{t}$ and $\alpha_{t}$ of Alternator++ need to be tuned for each application and each dataset, making them domain-specific. This can be time-consuming and may limit the application of Alternator++ to a narrower set of domains. Future work can consider adaptive noise scheduling techniques that dynamically adjust to varying noise levels within sequences, potentially improving performance on temporally heterogeneous data.

\bibliographystyle{apa}
\bibliography{arxiv}


\appendix

\section{Datasets Details}
\label{app:datasets:}
To test the Alternator++ on real datasets, We use the Monash time series repository \cite{godahewa2021monash}, which contains a group of 30 diverse real-world time-series datasets. From this repository, we specifically select the Solar Weekly, COVID death, FRED-MD, and NN5 Daily datasets as our focus of analysis and experimentation. These datasets have been chosen due to they reflect a variety of dynamics and challenges that allow us to thoroughly assess the capabilities of our model. We used the same setting for Alternator++ here as we used for the spiral dataset.

\parhead{Solar.} The Solar dataset represents the temporal aspects of solar power production within the United States during the year 2006. This specific sub-collection is dedicated to the state of Alabama and encompasses 137 individual time series, each delineating the weekly solar power production for a discrete region within the state during the aforementioned year\cite{lai2017modeling}. The temporal sequences encapsulated within this dataset effectively capture nuanced patterns, reflecting both seasonal fluctuations and geographical disparities in solar power generation across the United States. Therefore, the Solar dataset serves as a valuable resource for the evaluation and validation of generative models within the domain of time series analysis, with a specific emphasis on seasonal data dynamics.

\parhead{Covid.} The Covid dataset time series represents the fatalities for various countries and regions worldwide and was sourced from the Johns Hopkins University Center for Systems Science and Engineering (JHU CSSE). This dataset encompasses 266 daily time series, delineating the trajectory of COVID-19 fatalities across 43 distinct regions, comprising both states and countries. These temporal sequences within the dataset show trends and patterns in fatality rates across diverse regions. Consequently, the COVID-19 dataset assumes a pivotal role as a valuable resource for the examination and validation of generative models, particularly within the realm of time series analysis, with a specific emphasis on dynamic trends \cite{dong2020interactive}.

\parhead{Fred.} The Federal Reserve Economic Data (FRED) dataset represents an extensive and dynamic repository encompassing a diverse range of macroeconomic indicators, meticulously curated from the FRED-MD database\cite{mccracken2016fred}. This dataset is intentionally structured to facilitate recurring monthly updates, encapsulating 107 distinct time series spanning a duration of roughly 12 years (equivalent to 146 months). These time series eloquently depict various macroeconomic metrics, primarily procured from the Federal Reserve Bank.

\parhead{NN5.} The NN5 dataset shows daily ATM cash withdrawals in cities across the United Kingdom\cite{taieb2012review}. This dataset became well-known as a central part of the NN5 International Forecasting Competition, providing a deep look into the complex dynamics of ATM cash transactions in the banking field. There are 111 individual time series in this dataset, each covering around two years of daily cash withdrawal data, totaling 735 data points for each series. Its complexity comes from various time patterns, such as different seasonal cycles, local trends, and significant changes in how people withdraw cash over time. These features make it a valuable resource for testing and assessing generative models in the area of time series analysis.

\section{Implementation and Hyperparameter Search}
\label{app:imp}

We conducted comprehensive hyperparameter optimization and implementation refinements for Alternator++, carefully customized for each experimental task—density estimation, imputation, and forecasting—to maximize performance across all datasets. This section provides detailed insights into our implementation strategies, hyperparameter optimization approaches, and presents the results in the accompanying tables.

\paragraph{Density Estimation.} For density estimation tasks, we configured Alternator++ with a latent dimensionality ($D_z$) of 32, identified through exhaustive grid search across values of 16, 32, and 64. The architecture incorporates four specialized networks—$f_{\theta}^x(.)$, $f_{\phi}^z(.)$, $\boldsymbol{\epsilon}_z^{(b)}$, and $\boldsymbol{\epsilon}_x^{(b)}$—each constructed with two layers of self-attention mechanisms. Training utilized the Adam optimizer beginning with a learning rate of $1\times10^{-3}$, which gradually decreased to $1\times10^{-5}$ over 1000 epochs following a cosine annealing schedule. We processed data in batches of 100 samples. The noise variance parameters were determined through rigorous hyperparameter exploration, evaluating $\sigma_x$ and $\sigma_z$ across a range of values including 0.05, 0.1, 0.15, 0.2, and 0.3. Our experiments revealed optimal performance with $\sigma_x = 0.3$ and $\sigma_z = 0.15$ consistently across datasets. We employed a fixed, linearly spaced noise schedule throughout all density estimation experiments to ensure methodological consistency.

\begin{table}[h]
\centering
\caption{Hyperparameters for Density Estimation Experiments}
\begin{tabular}{lc}
\toprule
\textbf{Hyperparameter} & \textbf{Value} \\
\midrule
Latent Dimension ($D_z$) & 32 \\
Learning Rate & $1\times10^{-3}$ to $1\times10^{-5}$ (cosine annealing) \\
Batch Size & 100 \\
Noise Variances ($\sigma_x$, $\sigma_z$) & 0.3, 0.15 \\
Epochs & 1000 \\
Noise Schedule & Fixed (linearly spaced) \\
\bottomrule
\end{tabular}
\end{table}

\paragraph{Time-Series Imputation.} The imputation experiments leveraged the architectural foundation established in our density estimation setup, with modifications tailored to the unique challenges of handling missing data. Our hyperparameter optimization strategy prioritized developing robust performance across varying levels of data missingness, with particular emphasis on scenarios with high proportions of missing values. We trained the model using the Adam optimizer with an initial learning rate of $5\times10^{-4}$, gradually reducing to $5\times10^{-6}$ over 800 epochs through cosine annealing. To accommodate the increased variability inherent in imputation tasks, we employed a reduced batch size of 32 samples. Through systematic experimentation, we determined that noise variances of $\sigma_x = 0.15$ and $\sigma_z = 0.15$ provided optimal performance. Missing values were systematically introduced using a Missing At Random (MAR) protocol to simulate realistic data scenarios. Across all imputation tasks, we maintained a consistent approach with a fixed, linearly spaced noise schedule to ensure experimental rigor and comparability.

\begin{table}[h]
\centering
\caption{Hyperparameters for Time-Series Imputation Experiments}
\begin{tabular}{lc}
\toprule
\textbf{Hyperparameter} & \textbf{Value} \\
\midrule
Latent Dimension ($D_z$) & 64 \\
Learning Rate & $5\times10^{-4}$ to $5\times10^{-6}$ (cosine annealing) \\
Batch Size & 32 \\
Noise Variances ($\sigma_x$, $\sigma_z$) & 0.15, 0.15 \\
Epochs & 800 \\
Missing Data Rate & 10\% to 90\% (MAR) \\
Noise Schedule & Fixed (linearly spaced) \\
\bottomrule
\end{tabular}
\end{table}
\paragraph{Sea Surface Temperature Forecasting.}
For our Sea Surface Temperature (SST) forecasting task, we train two Adversarial Diffusion Models (ADM)~\citep{dhariwal2021diffusion}: one for the OTN and FTN components, and another for the scoring functions. Each model employs a U-Net backbone with attention modules placed after each CNN block. The backbone is configured with 128 base channels, 2 ResNet blocks per resolution, and a hierarchical channel multiplier structure of $\{1, 2, 4\}$ to capture complex spatial-temporal dynamics across multiple resolutions.

The models are trained for 700K iterations using a batch size of 8. We use the AdamW optimizer with parameters $\beta_1 = 0.9$, $\beta_2 = 0.999$, and an initial learning rate of $1 \times 10^{-4}$. For the diffusion process, noise schedules are calibrated with $\sigma_z = 0.1$, $\sigma_x = 0.2$, and $\alpha_t = 0.5$ for all time steps. These settings balance stochastic exploration with deterministic prediction, allowing the model to capture both short-term patterns and long-range uncertainties inherent in climate dynamics.

\begin{table}[!ht]
\centering
\caption{Model configuration for SST forecasting.}
\begin{tabular}{l|c}
\toprule
Number of ResNet blocks         & 2 \\
Base channels                   & 128 \\
Channel multipliers             & 1, 2, 4 \\
Attention resolutions           & 16 \\
Label dimensions                & 10 \\
Parameters (M)                  & 55.39 \\
\bottomrule
\end{tabular}
\label{tab:sst_model_config}
\end{table}

\begin{table}[!ht]
\centering
\caption{Training hyperparameters for SST forecasting.}
\begin{tabular}{l|c}
\toprule
Learning rate                   & $1 \times 10^{-4}$ \\
AdamW ($\beta_1$, $\beta_2$)    & (0.9, 0.999) \\
Batch size                      & 8 \\
Number of iterations            & 700K \\
GPU                             & NVIDIA A100 \\
\bottomrule
\end{tabular}
\label{tab:sst_hyperparams}
\end{table}

All SST experiments were conducted on NVIDIA A6000 GPUs with 48GB of memory, enabling efficient processing of the high-dimensional spatial-temporal inputs essential for accurate SST forecasting.

\end{document}